\documentclass{article}

\usepackage{times}
\usepackage{epsfig}
\usepackage{graphicx}
\usepackage{amsmath}
\usepackage{amssymb}
\usepackage{caption}
\usepackage{subcaption}
\usepackage{hyperref}
\usepackage{geometry}
\begin{document}

\title{Learning Good Representation via Continuous Attention}

\author{Liang Zhao\\
Baidu Research Institute\\
Sunnyvale, CA, USA\\
{\tt\small zhaoliang07@baidu.com}
\and
Wei Xu\thanks{This work was done when the author was at Baidu Research.}\\
Horizon Robotics Research\\
Sunnyvale, CA, USA\\
{\tt\small wie.xu@horizon.ai}
}

\date{\vspace{-5ex}}
 
\maketitle

\begin{abstract}
In this paper we present our scientific discovery that good representation can be learned via continuous attention during the interaction between Unsupervised Learning(UL) and Reinforcement Learning(RL) modules driven by intrinsic motivation. Specifically, we designed intrinsic rewards generated from UL modules for driving the RL agent to focus on objects for a period of time and to learn good representations of objects for later object recognition task. We evaluate our proposed algorithm in both with and without extrinsic reward settings. Experiments with end-to-end training in simulated environments with applications to few-shot object recognition demonstrated the effectiveness of the proposed algorithm.
\end{abstract}

\section{Introduction}
In Artificial Intelligence(AI), our ultimate goal is to develop intelligent machines and robots that could automatically acquire skills and knowledge under the guidance of intrinsic motivation \cite{schmidhuber2006:im} before given any specific tasks. The learned skills and knowledge will help improve the performance of downstream tasks
such as recognition, detection, navigation, and planning. In this paper, we focus on studying how to learn good representation or features of objects in a simulated 3D environment driven by intrinsic motivation. 

\begin{figure*}[h]
\begin{center}
\includegraphics[width= 0.8  \textwidth]{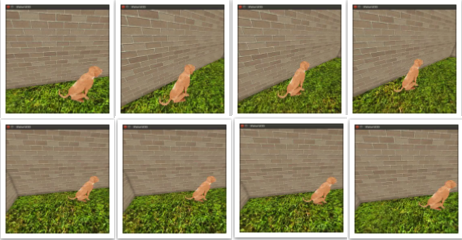}
\end{center}
\caption{Example of continuous attention on a dog by an RL agent driven by intrinsic motivation.}
\label{fig:att-dog}
\end{figure*}

Intrinsic motivation plays an important role in human development and learning, and researchers in many areas of cognitive science have emphasized that intrinsically motivated behavior is vital for intellectual growth. Inspired by human infants that exhibit strong intrinsic motivation and self-supervised capability to interact with the objects in the environment,
in this paper we propose a novel setting with deep neural networks for an agent to learn representations of objects in a 3D environment driven by the interaction between unsupervised learning and reinforcement learning modules. The RL module is responsible for collecting data from the environment and the UL module is responsible for extracting good features from the collected data and meanwhile for providing intrinsic rewards for the agent to explore the environment.  
The first skill for the agent to acquire is to learn a policy to attend to objects. Our observation is that random exploration cannot achieve frequent continuous object attention or continuous focusing on an object (as shown in Figure~\ref{fig:att-dog}) which is import for an agent to do online learning of the representation of the environment and objects.

This work contrasts with traditional passive unsupervised learning. Instead of learning representation of the given input, the agent actively choose attention direction and training data for learning meaningful/informative features as demonstrated in the experiment section.
We propose an alternative to unsupervised learning that takes advantage of active control where to collect data in the environment via an exploration policy.

The main contribution is our scientific discovery that good representation can be learned via continuous attention during the interaction between Unsupervised Learning(UL) and Reinforcement Learning(RL) modules driven by intrinsic motivation. Specifically, we designed intrinsic rewards generated from UL modules for driving the RL agent to focus on objects for a period of time and to learn good representations of objects for later object recognition task. 

We evaluate our proposed algorithm in both with and without extrinsic reward settings. Experiments with end-to-end training in simulated environments with applications to few-shot object recognition demonstrated the effectiveness of the proposed algorithm.

The rest of the paper is organized as follows. We briefly review related work in Section 2. The detailed description of the proposed deep neural network architecture and algorithm are given in Sections 3 and 4, respectively, followed by experimental results in Section 5 and conclusion in Section 6.

\section{Related Works}
Our work is closely related to intrinsic motivation, representation learning and their application to few-shot object recognition. Here, we briefly review some related work in these areas.

\textbf{Representation Learning: }
Learning good data representations (or features) is one of the main goals for any machine learning algorithms --no matter whether it is a supervised or unsupervised one. For unsupervised learning, we want the learned representations of the data makes it easier to build classifiers or other predictors upon them. The question is how to evaluate the quality of learned representations. Goodfellow et al. \cite{goodfellow2009:ftr} proposed a number of empirical tests for directly measuring the degree to which learned representations are invariant to different transformations. Higgins et al. \cite{higgins2016:ul} devised a protocol to quantify the degree of disentanglement learned by different models. Bengio et al. \cite{bengio2013:ftr} list some properties/attributes a good representation should possess, such as sparsity, distributed among multiple explanatory factors, hierarchical organization with more abstract and invariant concepts \cite{hubel1962:ftr, wiskott2002:ftr} higher in the hierarchy, temporal and spatial coherence. The listed properties provide us guide to design appropriate deep neural network architecture and algorithm for learning such representations.

\textbf{Intrinsic Motivation: }
The goal of RL agents is to maximize discounted cumulative reward
\cite{sutton2016:rl, mnih2015:dqn, mnih2016:a3c}.
Reward functions should be defined generically and lead to good long-term outcomes for an agent. However, most RL algorithms suffer from sparse rewards. Reward shaping \cite{ng1999:policyinvariance} provides a way to deal with sparse extrinsic rewards; anther way is intrinsic motivation \cite{schmidhuber2006:im}. 
Intrinsic rewards such as prediction gain \cite{bellemare2016:im}, learning progress \cite{oudeyer2007:im}, compression progress \cite{schmidhuber2010:im}, variational information maximization \cite{abbeel2016:vime, hester2017:im}, have been employed to augment the environment's reward signal for encouraging to discover novel behavior patterns.

\textbf{UL+RL: }
Combining UL and RL modules have been proposed in previous work. Most
 work \cite{stadie2015:predict, jaderberg2017:rlul, pathak2017:rlul} focus on improving reinforcement learning performance using unsupervised learning tasks to achieve sample efficiency with higher return. In contrast, our work focus on learning good representation of objects via continuous attention driven by interaction between UL and RL modules. The learned RL exploration policy helps the UL module learn better representation (features) of objects by providing informative data to the UL module.

\textbf{Attention: } Attention mechanism has been proposed to help visualize, interpret the inner working of deep neural networks and to improve the performance of many machine learning tasks such as machine translation \cite{bahdanau2014:attn, vaswani2017:attn},
 image classification \cite{mnih2014:attn, cao2015:attn, jaderberg2015:attn, jetley2018:attn}, visual question answering \cite{xu2016:attn, yang2016:attn, andreas2016:attn},  and image captioning \cite{xu2015:attn, you2016:attn, mun2017:attn}. 
Unlike these previous work whose goal is to estimate a 2D spatial or 1D temporal attention map, our work focus on training an agent to attend to objects in a 3D environment for a period of time in order to learn a good representation of the object.

\section{Deep Neural Network Architecture}
\begin{figure}[h]
\begin{center}
\begin{subfigure}[b]{0.4 \textwidth}
\includegraphics[width=\textwidth]{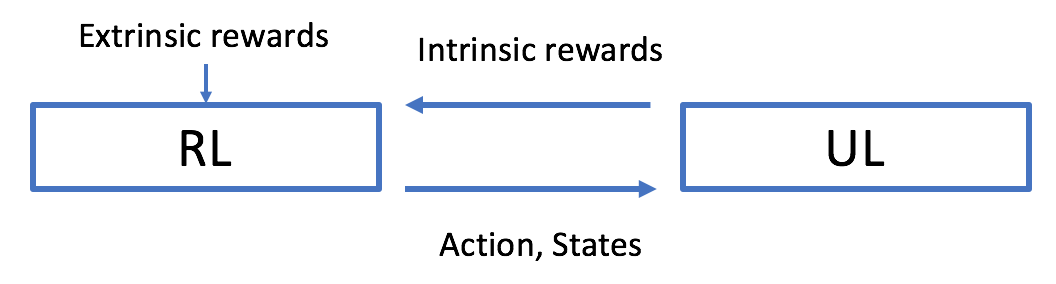}
\caption{\label{sfig:rl-sparse}}
\end{subfigure}
\begin{subfigure}[b]{0.4 \textwidth}
\includegraphics[width=\textwidth]{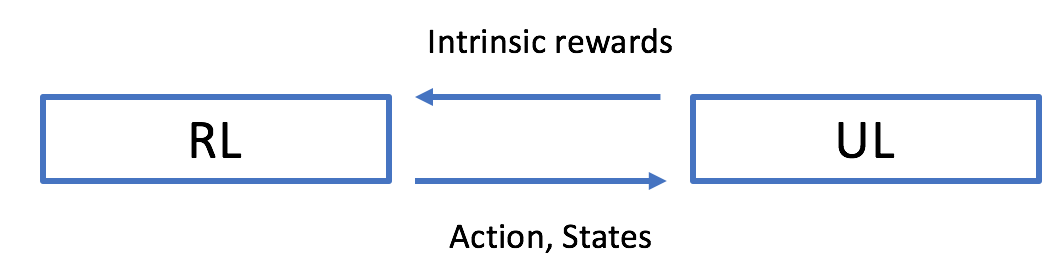}
\caption{\label{sfig:rl-ul}}
\end{subfigure}
\end{center}
\caption{General framework: (a) with extrinsic rewards (b) without extrinsic rewards}
\label{fig:framework}
\end{figure}

See Figure~\ref{fig:framework} for the general framework which includes a RL and a UL modules: UL module provides intrinsic rewards to the RL module while RL module provides training data to the UL module.

\textbf{RL module architecture: } The RL module in Figure~\ref{fig:framework} follows the A3C architecture \cite{mnih2016:a3c}. First, the input state $s_t$ is passed through a sequence of four convolution layers with 32 filters each, kernel size of 3x3, stride of 2 and padding of 1. A RELU activation unit is used after each convolution layer. Then, two separate fully connected layers are used to predict the value function and the action from the feature representation of the forth convolution layer.

\begin{figure*}[h]
\begin{center}
\includegraphics[width= 0.8  \textwidth]{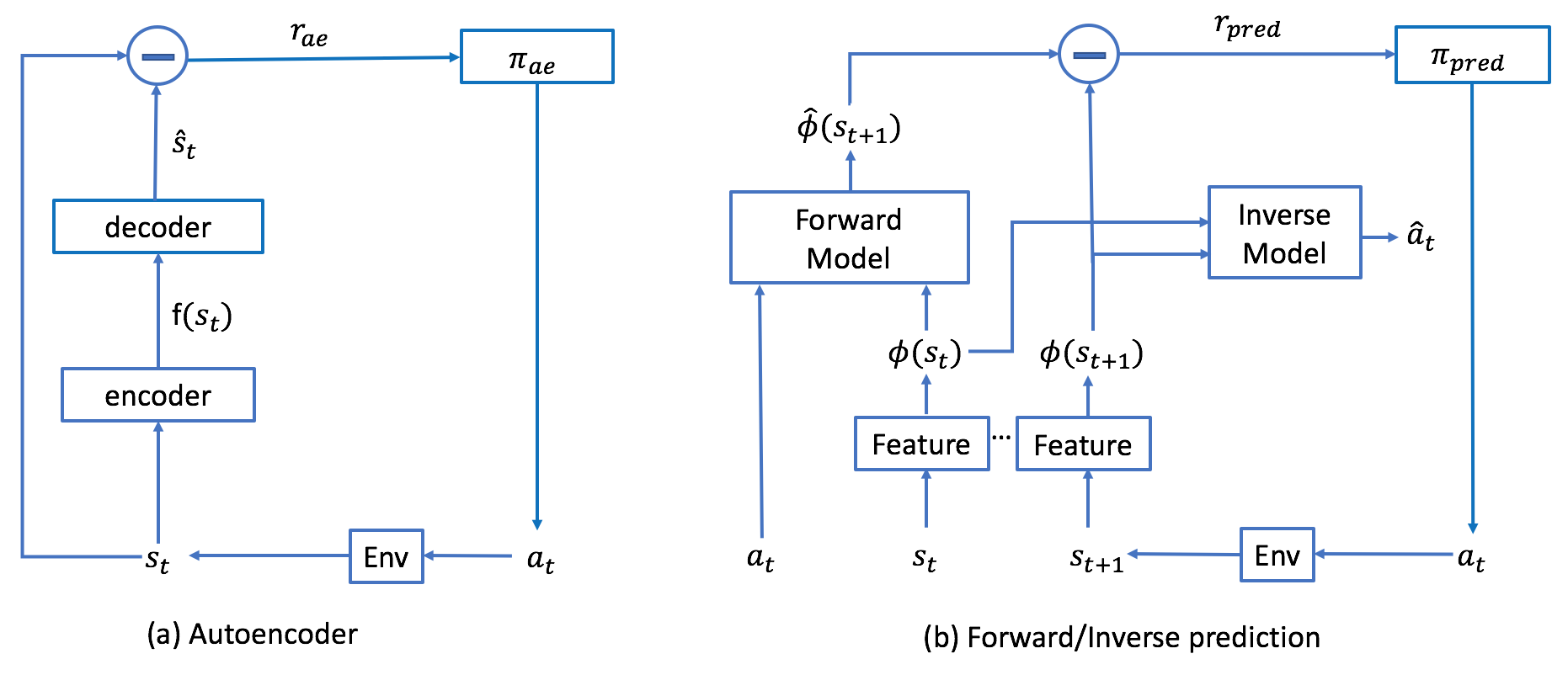}
\end{center}
\caption{UL module architectures: (a) autoencoder (b) prediction}
\label{fig:arch}
\end{figure*}

Figure~\ref{fig:arch} shows two designs of the UL module: autoencoder and prediction modules.

\textbf{UL autoencoder architecture: } In Figure~\ref{fig:arch}(a), the encoder is a sequence of three convolution layers with 16, 32, 64 filters, respectively, kernel size of 4x4, stride of 4 and padding of 0. The decoder is a sequence of three deconvolution layers corresponding to three convolution layers in the encoder. The reconstruction error is used as intrinsic rewards to the RL module.

\textbf{UL prediction architecture: } As shown in Figure~\ref{fig:arch}(b), the prediction module follows the design in \cite{pathak2017:rlul}. It contains a dynamic forward model and a dynamic inverse model. The dynamic forward model predicts the feature of next state given the feature of the current state and agent's action and the prediction error is used as intrinsic reward to the RL module. The dynamic inverse model predicts the agent action that causes the transition of current state to the next state. The feature is extracted by a sequence of three convolution layers with 16, 32, 64 filters, respectively; kernel size of 6x6, 3x3, 3x3, respectively; stride of 6, 3, 3, respectively; and padding of 0.

The autoencoder focuses on learning the static features of the scene and provides the signal to make the learned representation compressed and meanwhile containing enough information to reconstruct the scene. This is complement to the signal provided by the forward dynamics model which focuses on encoding the dynamic aspect of the scene.

\section{Algorithm}
The algorithm includes two phases. The goal of phase one is to learn representation of objects in unsupervised setting via interaction between UL and RL modules; while the goal of phase two is to
use the learned representation to improve the performance of object recognition in supervised setting. Self-designed task in phase one is to learn object representation from different points of view. Recognition training stage: provide label for image of an object from a few points of views;
Recognition test stage: test recognition of an object from the rest points of views.

\subsection{Intrinsic Rewards Generated by UL Modules}
The core of learning (both for human and machine) is to learn from mistake and to learn from failure. We use the following two loss functions as intrinsic rewards for autoencoder and prediction modules, respectively. These intrinsic rewards drive the agent to explore areas where there are large reconstruction and prediction errors and therefore provide the UL module more diverse and informative data. 

\begin{equation}
r_{ae} = L_{ae} = || s_t - \hat{s}_t ||_2
\end{equation}

\begin{equation}
r_{pred} = L_{pred} =  || \phi(s_{t+1}) - \hat{\phi}(s_{t+1}) ||_2
\end{equation}

\section{Experiments}
\subsection{Experimental Setting}
\begin{figure*}[th]
\begin{center}
\begin{subfigure}[b]{0.4 \textwidth}
\includegraphics[width=\textwidth]{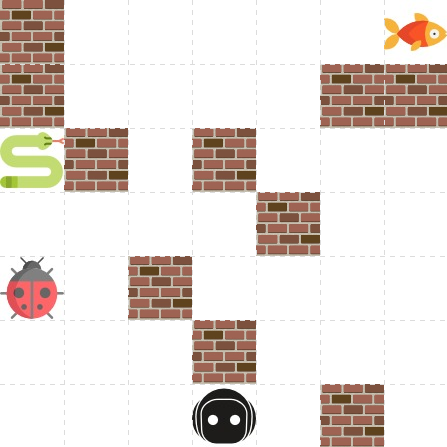}
\caption{\label{sfig:xw2d} }
\end{subfigure}
\begin{subfigure}[b]{0.4 \textwidth}
\includegraphics[width=\textwidth]{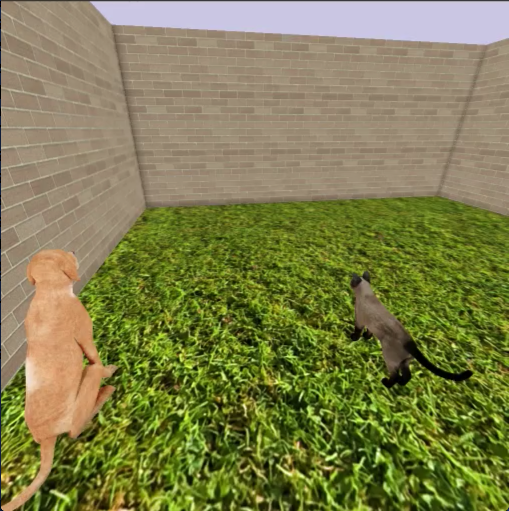}
\caption{\label{sfig:xw3d} }
\end{subfigure}
\end{center}
\caption{Snapshots of (a) Xworld2D (b) XWorld3D}
\label{fig:xw}
\end{figure*}

\textbf{Environments} The first environment we evaluate on is XWorld2D \footnote{https://github.com/PaddlePaddle/XWorld/games/xworld} with navigation task which is used to demonstrate that auxiliary unsupervised learning modules improve policy learning when there are only sparse extrinsic rewards. Figure~\ref{fig:xw}(a) illustrates a snapshot of XWorld2D, which is a 2D world for an agent to learn vision and language skills. In this paper, we train the agent on navigation tasks given by a virtual teacher. The goal is given by a word such as ``fish". The agent should learn simultaneously the visual representation of the object and the action control that moves the agent to the goal. The agent gets a reward only when it reaches the goal. Therefore, this is a typical sparse reward reinforcement learning problem.

The second one is a simulated 3D environment -- XWorld3D \footnote{https://github.com/PaddlePaddle/XWorld/games/xworld3d} which is used to demonstrate that the learned representation via interaction between UL and RL modules helps improve the performance of downstream few-shot learning task.
Figure~\ref{fig:xw}(b) illustrates a snapshot of XWorld3D. In this paper, we train the agent to explore the 3D world without given it any tasks in phase one. The agent sees the world in a first-person view, learns simultaneously the visual representations of the objects, and how to explore the world driven by intrinsic motivation. Therefore, this is a typical reinforcement learning problem without extrinsic reward which mimics how a baby learns a model of the environment without any guide.

\textbf{Training details} All agents are trained using visual inputs. The input RGB images are resized to 64x64. Following the asynchronous training protocol in A3C, all the agents were trained asynchronously with sixteen workers using stochastic descent. We used RMSprop \cite{tieleman2012:opt} optimizer with a learning rate of $10^{-5}$ , a damping factor of 0.01, and a gradient moving average decay of 0.95. The gradient has a momentum of 0.9. The batch size is set to 32. The training details described in this section apply to all the methods in both environments.

\textbf{Baseline method} ``a3c + ae" and ``a3c + pred" denote two ``UL+RL" algorithms. We compare our algorithms with a baseline method where A3C module is replaced with a random policy which generates actions by uniformly sampling from a set of actions (move\_forward, move\_backward, move\_left, move\_right, turn\_left, and turn\_right).

\subsection{Results with Extrinsic Rewards}
Figure~\ref{fig:xw}(a) shows the XWorld2D simulation environment which we used to conduct experiments with extrinsic reward.

\begin{figure}[h]
\centering
\includegraphics[width=0.8\textwidth]{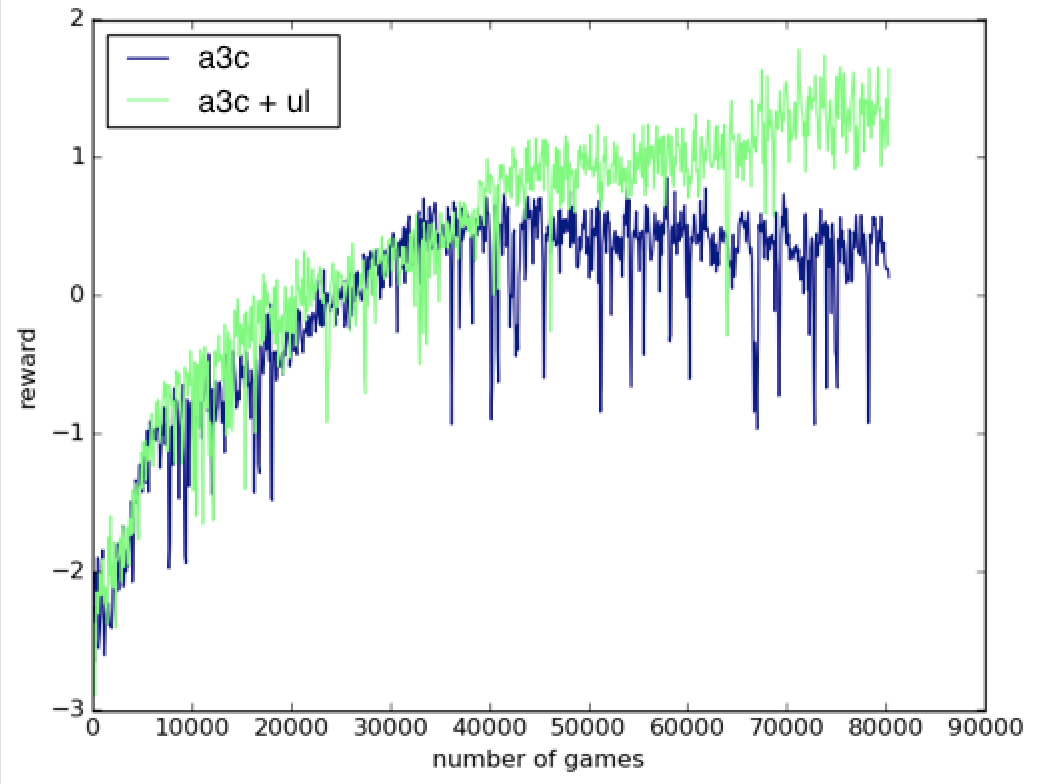}
\caption{Average return on XWorld2D navigation tasks}
\label{fig:xw2d-rw} 
\end{figure}

Figure~\ref{fig:xw2d-rw} illustrates the average rewards obtained by the agent in XWorld2d navigation tasks with and without intrinsic rewards. The results demonstrate that auxiliary UL module improve RL policy learning for achieving higher average reward when there are only sparse extrinsic rewards.

\subsection{Interaction between UL and RL Modules}
\begin{figure*}[th]
\begin{center}
\begin{subfigure}[b]{0.3 \textwidth}
\includegraphics[width=\textwidth]{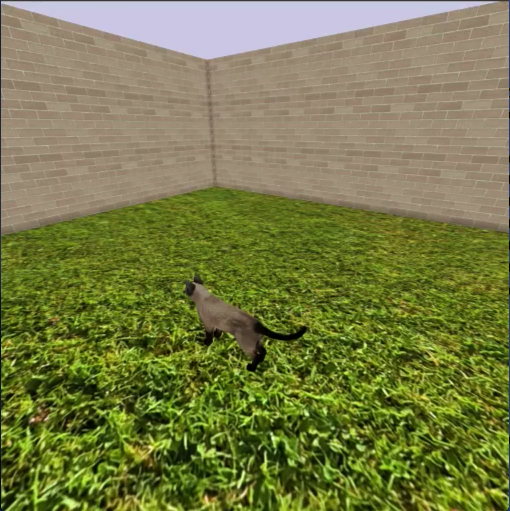}
\caption{\label{sfig:xw3d-1} }
\end{subfigure}
\begin{subfigure}[b]{0.3 \textwidth}
\includegraphics[width=\textwidth]{xworld3d-2.png}
\caption{\label{sfig:xw3d-2} }
\end{subfigure}
\begin{subfigure}[b]{0.3 \textwidth}
\includegraphics[width=\textwidth]{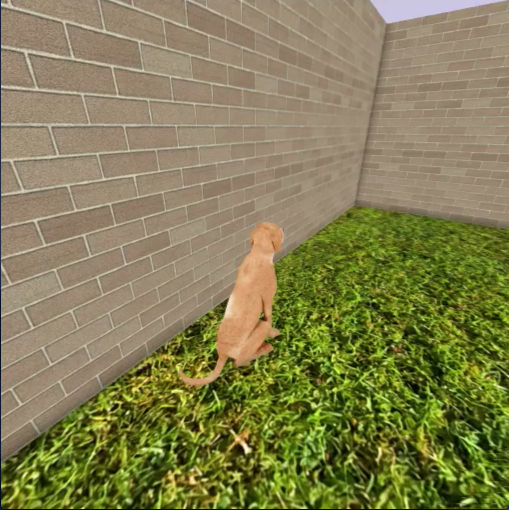}
\caption{\label{sfig:xw3d-3} }
\end{subfigure}
\end{center}
\caption{Snapshots of XWorld3D}
\label{fig:xw3d}
\end{figure*}

Figure~\ref{fig:xw3d} shows the XWorld3D simulation environment which we used to conduct experiments without extrinsic reward.

\begin{figure*}[tb]
\begin{center}
\begin{subfigure}[b]{0.3 \textwidth}
\includegraphics[width=\textwidth]{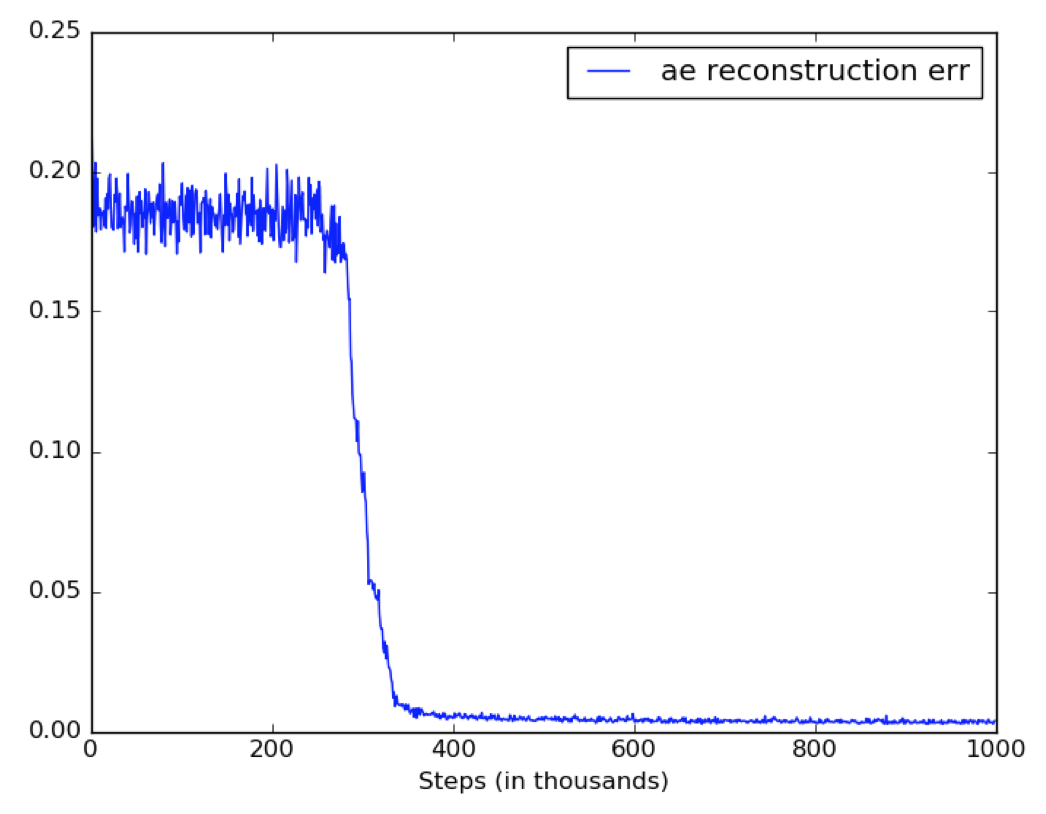}
\caption{\label{sfig:ae} } 
\end{subfigure}
\begin{subfigure}[b]{0.3 \textwidth}
\includegraphics[width=\textwidth]{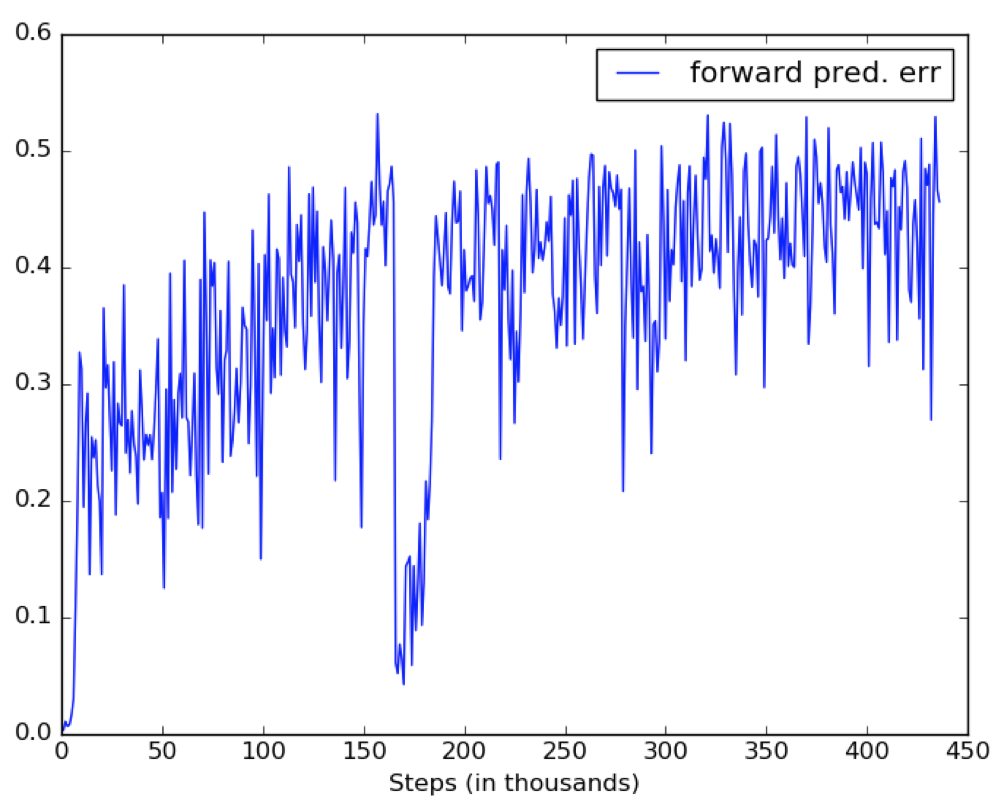}
\caption{\label{sfig:pred} } 
\end{subfigure}
\begin{subfigure}[b]{0.3 \textwidth}
\includegraphics[width=\textwidth]{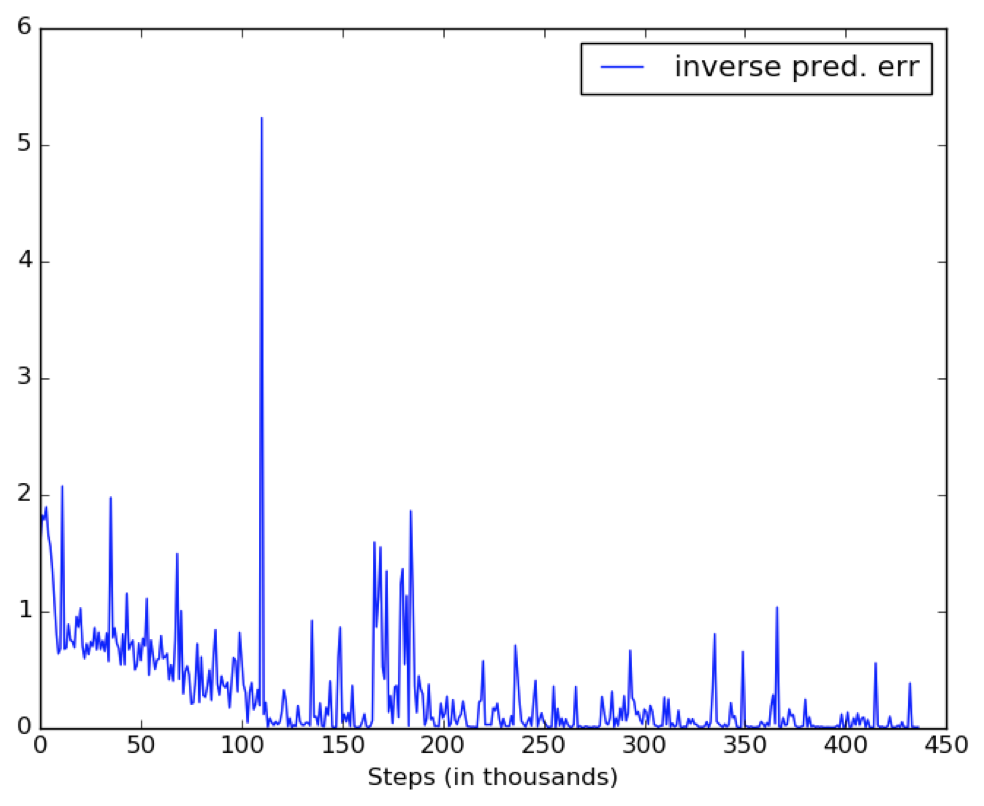}
\caption{\label{sfig:pred} } 
\end{subfigure}
\end{center}
\caption{(a) autoencoder reconstruction error (b) forward prediction error (c) inverse prediction error}
\label{fig:ir}
\end{figure*}

Figure~\ref{fig:ir} illustrates how the intrinsic rewards or unsupervised learning task performance changes as the reinforcement learning policy improves. For autoencoder (Figure~\ref{fig:ir}(a)), the reconstruction error is large at first and then drops to a low level and stays low, which means that the RL exploration policy samples data that cover the whole environment for training the autoencoder so that it learns how to encode and decode the state of the environment in an efficient way. 

For dynamic forward module (Figure~\ref{fig:ir}(b)), the agent first approach one object (a cat) as shown in Figure~\ref{fig:xw3d}(a) and learns how to predict the feature of the next state and the prediction error drops once the forward module learns how to predict the state of the object (the cat) in Figure~\ref{fig:xw3d}(a). The intrinsic reward drives the agent to approach to another new object (a dog) in the environment (Figure~\ref{fig:xw3d}(b)(c)) and makes the prediction error increase again as shown in Figure~\ref{fig:ir}(b). 

Dynamic inverse prediction is a relatively easier task comparing with dynamic forward prediction -- it only needs to learn how to predict self-motion given the features of the previous and current states. Therefore, once the module learns how to do it the inverse prediction error drops and stays at a low level as shown in Figure~\ref{fig:ir}(c).

\subsection{Object Attention Using Different RL Policies}
The first skill for the agent to acquire is to learn a policy to attend to objects. Our observation is that random exploration cannot achieve continuous and frequent object attention. See Table~\ref{tab:attn} for a list of the mean and standard deviation of object attention frequencies and durations for three policies -- random exploration, autoencoder, and prediction modules. The object attention frequency is calculated as the ratio between the total number of frames with an object present and the total number of frames within a time window; the object attention duration is calculated as the number of consecutive frames with the same object present. Table~\ref{tab:attn} demonstrates that the policies driven by both UL modules achieves higher object attention frequency (51\%) and longer continuous focus on the same object (9.6 and 12.1 frames, respectively) than random exploration policy does (frequency: 48\% and duration: 7.1 frames).

\begin{table}[ht]
\caption{Results on object attention task}
\label{tab:attn}
\begin{center}
\begin{tabular}{|c|c|c|c|}
\hline
RL Policy  & Random & Autoencoder & Prediction
\\ \hline
Frequency (\%) & 48$\pm$6 & 51$\pm$10 &\textbf{51$\pm$8}
\\ \hline
Duration & 7.1 $\pm$ 4.4 & 9.6 $\pm$ 5.4 &\textbf{12.1 $\pm$ 11.8}
\\ \hline
\end{tabular}
\end{center}
\end{table}

As shown in Fig~\ref{fig:ir}(b), the error increase corresponds to emergence of object attention. Both autoencoder and prediction modules exhibit increased frequency of consecutive frames with an object present coinciding with the increase 
in intrinsic rewards or reconstruction and forward prediction errors, respective
ly.

The RL agents driven by UL modules tend to attend to where they are going and more informative directions, for example they look straight ahead to an object when moving forward while agents trained with random exploration sometimes look in less informative directions (e.g. looking at the empty scene without any object 
present).

\subsection{Qualitative Evaluation of Learned Features}
\begin{figure}[h]
\begin{center}
\begin{subfigure}[b]{0.4 \textwidth}
\includegraphics[width=\textwidth]{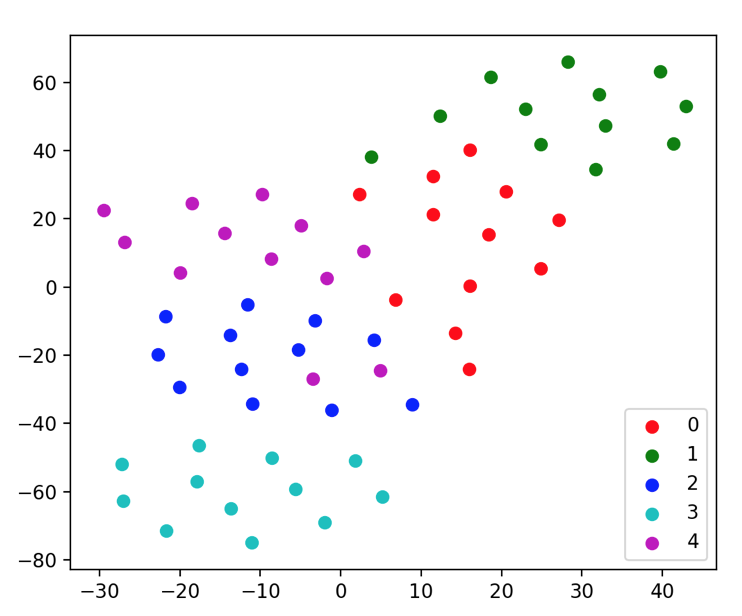}
\caption{\label{sfig:ae-ftr} AE features}
\end{subfigure}
\begin{subfigure}[b]{0.4 \textwidth}
\includegraphics[width=\textwidth]{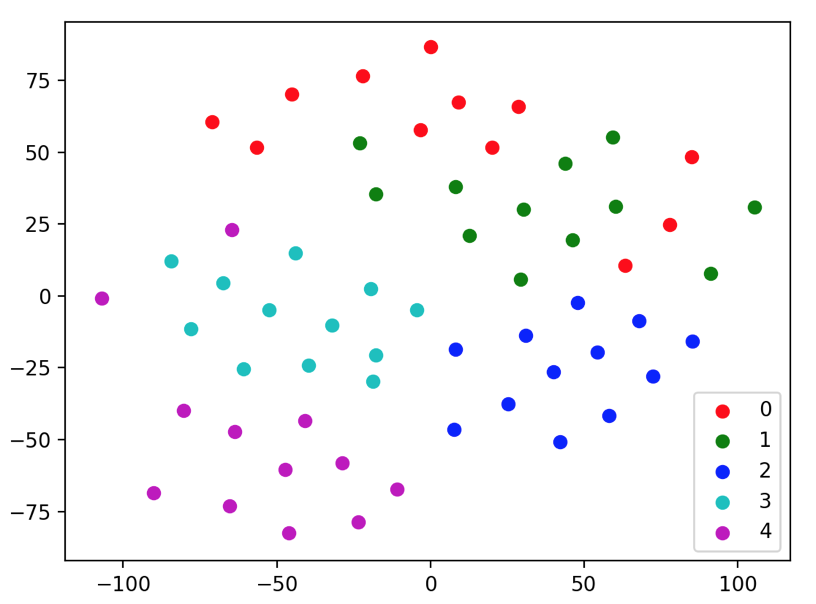}
\caption{\label{sfig:pred-ftr} Prediction features}
\end{subfigure}
\end{center}
\caption{t-SNE visualization of features learned from UL modules: (a) Autoencoder (b) Prediction}
\label{fig:ftr}
\end{figure}

We qualitatively evaluate the arrangement of learned features using t-SNE representations \cite{maaten2008:tsne}. Figure~\ref{fig:ftr} visualize the features learned in phase one by autoencoder and prediction modules, respectively. It demonstrates that both UL modules can learn discriminative features for grouping objects of the same category into the same cluster.

\subsection{Quantitative Evaluation of Learned Features}
\begin{figure*}[htbp]
\begin{center}
\begin{subfigure}[b]{0.2 \textwidth}
\includegraphics[width=\textwidth]{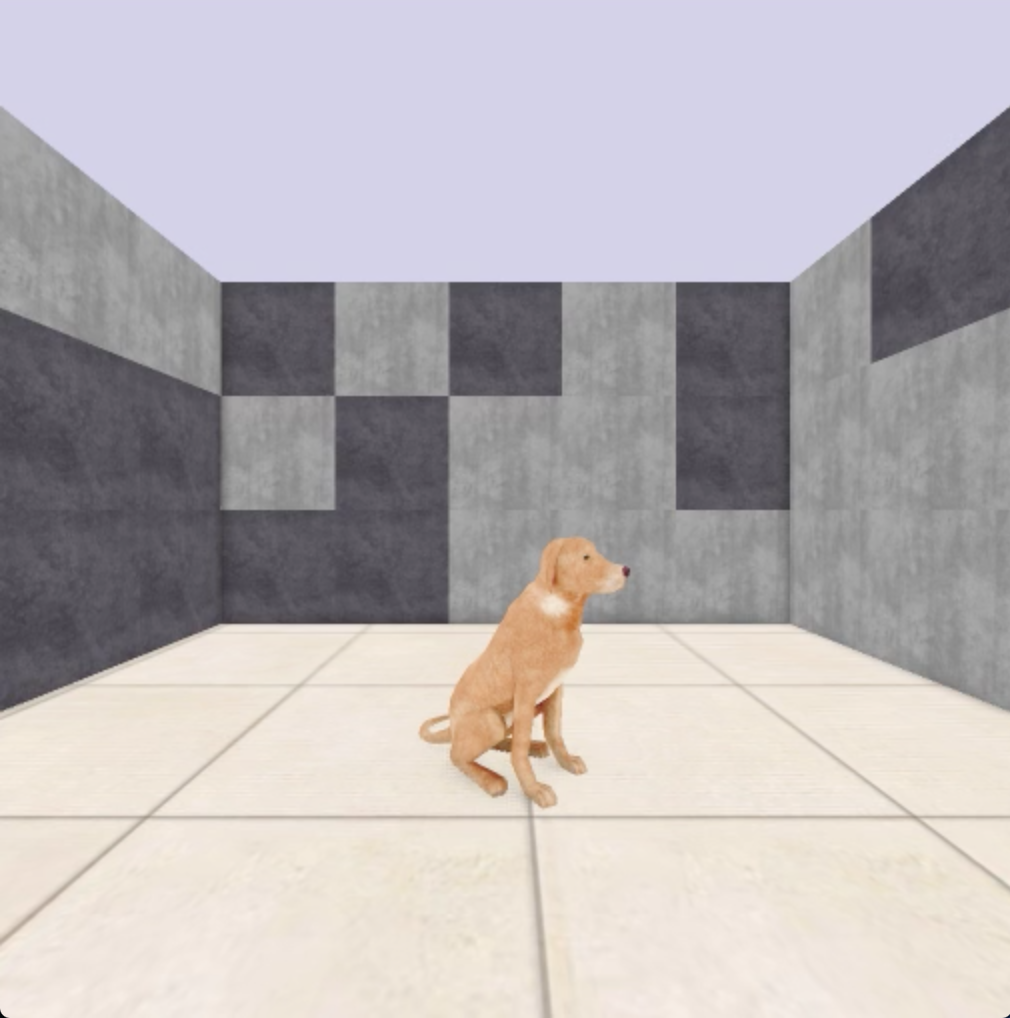}
\caption{\label{sfig:dog-1} }
\end{subfigure}
\begin{subfigure}[b]{0.2 \textwidth}
\includegraphics[width=\textwidth]{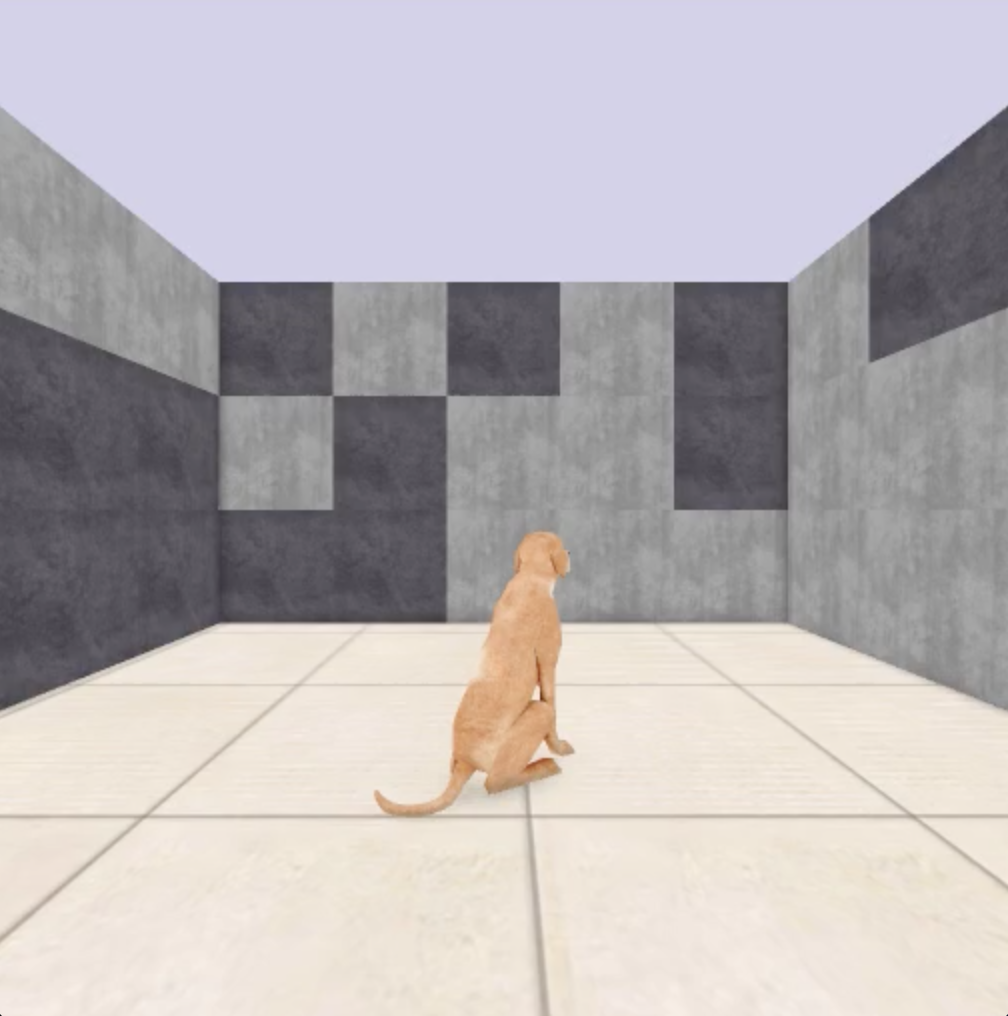}
\caption{\label{sfig:dog-2} }
\end{subfigure}
\begin{subfigure}[b]{0.2 \textwidth}
\includegraphics[width=\textwidth]{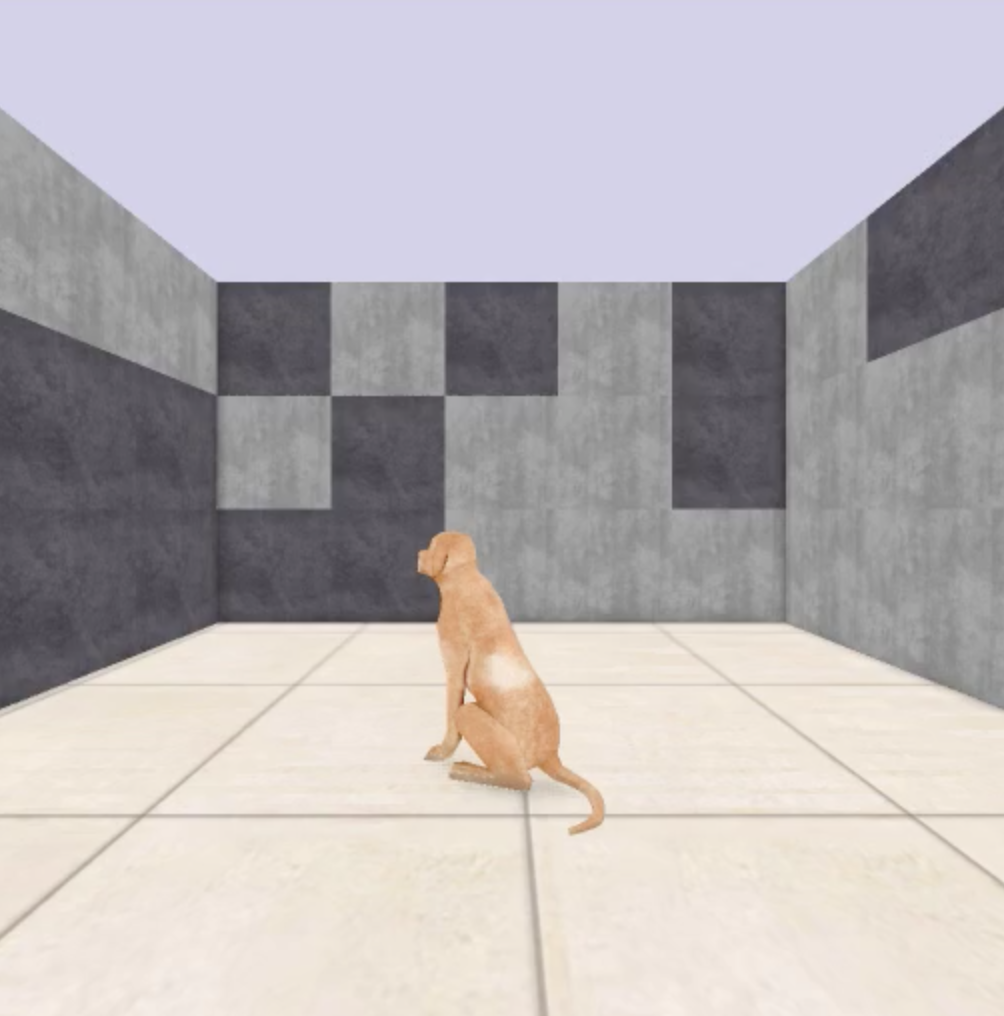}
\caption{\label{sfig:dog-3} }
\end{subfigure}
\begin{subfigure}[b]{0.2 \textwidth}
\includegraphics[width=\textwidth]{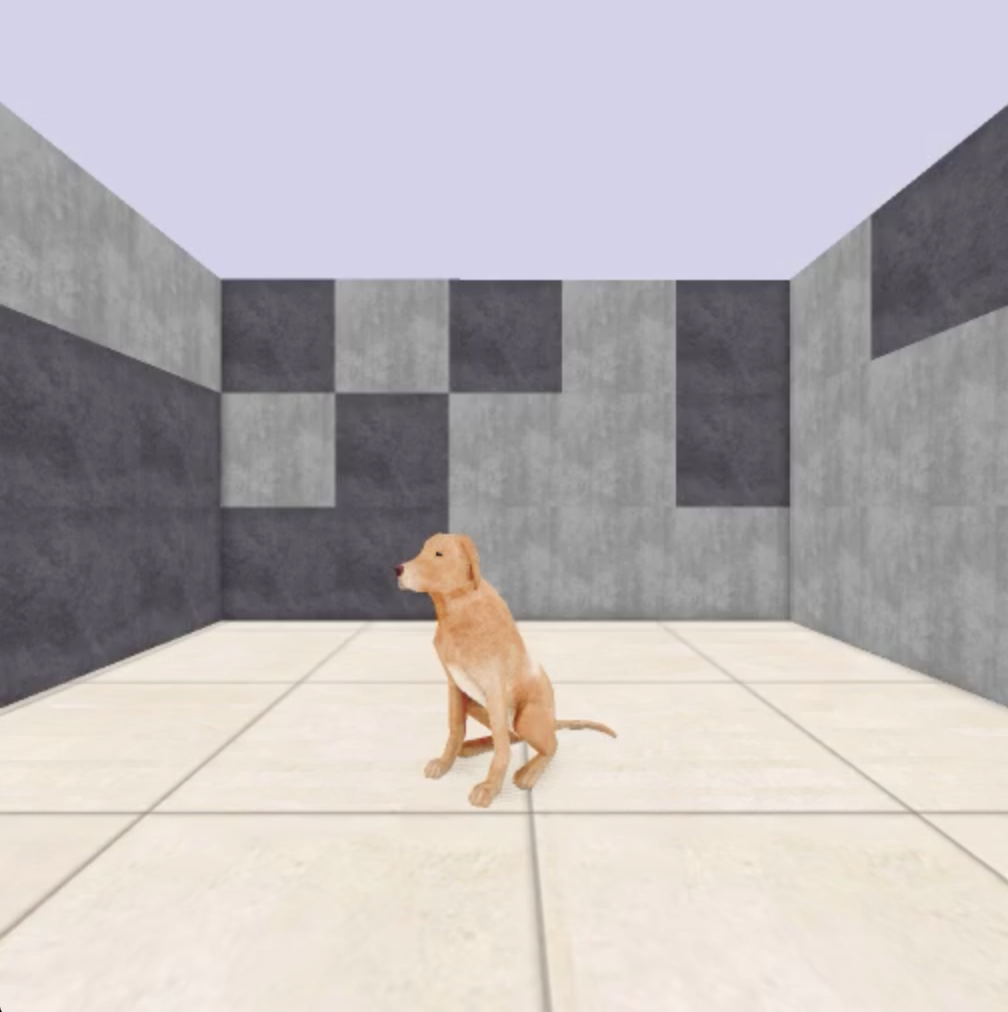}
\caption{\label{sfig:dog-4} }
\end{subfigure}
\end{center}
\caption{Snapshots of the XWorld3D for object recognition}
\label{fig:dog}
\end{figure*}

Figure~\ref{fig:dog} shows some snapshots of the XWorld3D environment for object recognition task. In the training stage, we provide labels for images of an object from a few randomly sampled points of views, while in the test stage, we evaluate recognition performance on the object from the rest points of views.

\begin{table}[ht]
\caption{Object recognition errors by AE and prediction modules}
\label{tab:sl}
\begin{center}
\begin{tabular}{|c|c|c|c|c|c|c|}
\hline
5-way-n-shot & 1 & 3 & 5 & 7
\\ \hline
AE & 27.0\% & 11.3\% & 11.0\% & {\bf 7.0\%}
\\ \hline
Prediction & 36.0\% & 28.9\% & 18.2\% & {\bf 14.0\%}
\\ \hline
\end{tabular}
\end{center}
\end{table}

As a proof of concept of the idea, Table~\ref{tab:sl} presents the object recognition results by two UL modules using k-nearest neighbors (KNN). The recognition errors decrease as the training data increase and achieves 7.0\% and 14.0\% for AE and prediction module, respectively, when the number of training data gets 7.
 The reason that AE module achieves lower recognition error than prediction module is probably that AE module converge to a lower reconstruction error than prediction error by prediction module within the same number of iterations (450k) as demonstrated in Fig~\ref{fig:ir}(a)(b).

\section{Conclusion and Future Work}
In this paper we present our scientific discovery that good representation can be learned via continuous attention during the interaction between Unsupervised Learning and Reinforcement Learning modules driven by intrinsic motivation. Specifically, we designed intrinsic rewards generated from UL modules for driving the RL agent to focus on objects for a period of time and to learn good representations of objects for later object recognition task. We evaluate our proposed algorithm in both with and without extrinsic reward settings. Experiments with end-to-end training in simulated environments with applications to few-shot object recognition demonstrated the effectiveness of the proposed algorithm. From the experiments we observe  that the policies driven by both UL modules achieves higher object attention frequency and longer continuous focus on the same object than random exploration policy does.  Both UL modules can learn discriminative features for grouping objects of the same category into the same cluster.

Several parts of the presented approach could be tuned and studied in more detail. The proposed deep neural network architecture could be used in a semi-supervised setting. The DNN could be extended to incorporate other UL tasks. We leave these questions for future research. We consider the present work as initial steps on the way to making an agent perform unsupervised learning in a real environment.




\end{document}